\documentclass{article}

\usepackage[preprint]{neurips_2026}

\usepackage[utf8]{inputenc} % allow utf-8 input
\usepackage[T1]{fontenc}    % use 8-bit T1 fonts
\usepackage{hyperref}       % hyperlinks
\usepackage{url}            % simple URL typesetting
\usepackage{booktabs}       % professional-quality tables
\usepackage{amsfonts}       % blackboard math symbols
\usepackage{nicefrac}       % compact symbols for 1/2, etc.
\usepackage{microtype}      % microtypography
\usepackage{xcolor}         % colors
\usepackage{graphicx}
\usepackage{amsmath,amssymb}
\usepackage{booktabs}
\usepackage{xcolor}
\usepackage{colortbl}
\usepackage{array}
\usepackage{multirow}
\usepackage{hyperref}
\usepackage[capitalise,noabbrev]{cleveref}
\usepackage{geometry}
\definecolor{tableheader}{RGB}{46,134,171}
\definecolor{tablerowalt}{RGB}{245,248,250}
\definecolor{bestresult}{RGB}{46,134,171}

\title{EVLA: An Electro-Aware Multimodal Assistant for Physically-Grounded Driving Reasoning and Control}

\author{%
  Yuxin Liu \quad Zihan Chen \quad Haoyu Wang \quad Mingxuan Zhang \\
  Ruijie Lin \quad Siyuan Zhao \\
  College of Computer Science and Technology\\
  Zhejiang University\\
  Hangzhou, Zhejiang, China
}

\begin{document}

\maketitle

\begin{abstract}
Modern vision-language models (VLMs) for driving assistants typically treat vehicle dynamics as a black box, resulting in decisions that lack awareness of the vehicle's real-time electro-mechanical state. To bridge this gap, we introduce the Electro-Visual-Language Assistant (EVLA)---a novel framework that combines multi-modal scene understanding with real-time perception of the electrified powertrain state (e.g., motor torque, battery SOC). Our approach features two key innovations: first, a Unified Co-State Encoder (UCSE) that fuses visual, textual, and vehicle-state inputs into a shared latent representation, augmented with an Energy-Efficiency Field to model spatial energy costs; and second, an Electro-aware Structured Reasoning Chain (ESRC), which replaces external chain-of-thought prompting with an internal, deterministic reasoning process grounded in physical constraints and optimization objectives. Trained end-to-end with a physics-guided joint loss, EVLA learns to generate context-aware and energy-optimal driving decisions. Extensive evaluations on a driving QA benchmark demonstrate that EVLA substantially outperforms strong fine-tuned VLM baselines, improving the final score by +0.0871 and accuracy by +5.6\%. Ablation studies validate the necessity of each component, and efficiency analyses show that EVLA achieves 36\% faster inference than multi-stage pipelines. This work underscores that integrating vehicle-state awareness and structured physical reasoning is crucial for developing next-generation, physically-grounded driving assistants.
\end{abstract}

\begin{figure}
    \centering
    \includegraphics[width=1\linewidth]{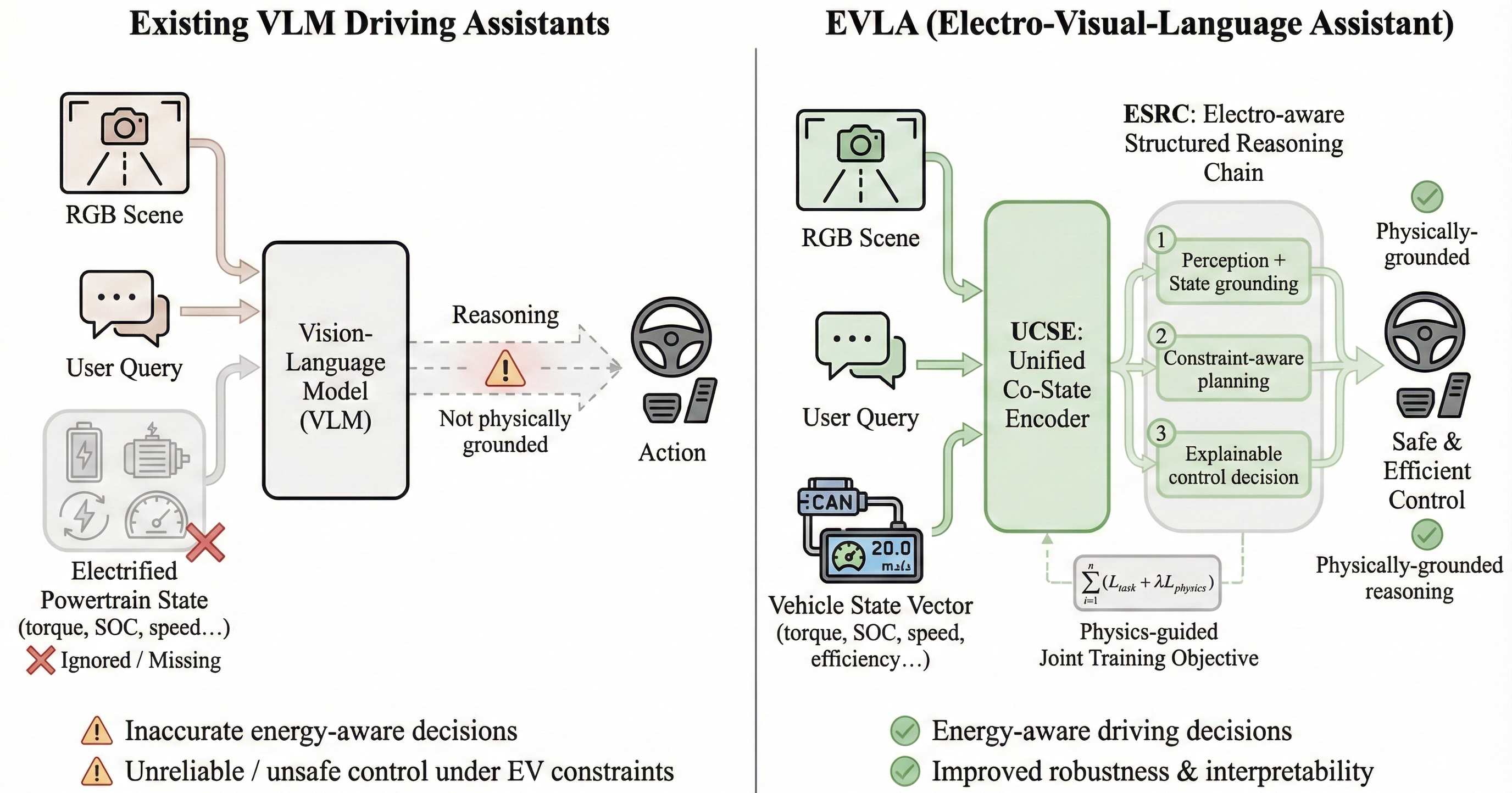}
    \caption{Motivation of EVLA. Existing vision--language driving assistants ignore electrified powertrain states, leading to physically ungrounded reasoning and unreliable control. EVLA explicitly integrates visual perception, language instructions, and vehicle state information to enable energy-aware and physically grounded driving decisions.}
    \label{fig:motivation}
\end{figure}

\section{Introduction}
\label{sec:introduction}

Recent advancements in Vision-Language Models (VLMs) have demonstrated considerable promise for intelligent driving systems. By jointly processing visual scenes and natural language, VLMs can interpret road conditions, detect obstacles, and answer complex queries about the driving environment, thereby enhancing the reasoning capabilities and situational awareness of autonomous agents~\cite{vlm_ad_survey, drivegpt4, llava, zhou2025reagent, qi2022capacitive}. This progress is exemplified by benchmarks such as the CVPR 2024 Driving with Language challenge, which focuses on developing models capable of addressing diverse driving questions using multi-view image inputs.

Despite these advances, a fundamental limitation persists. Existing VLM-based approaches for driving largely operate as passive visual question-answering systems. They treat the autonomous vehicle as a black box, lacking explicit comprehension of its internal \textit{electro-mechanical state}---including motor torque, battery state-of-charge, or thermal limits. This oversight hinders holistic reasoning for tasks such as energy-efficient planning, where decisions must integrate both external scene semantics and internal vehicle dynamics~\cite{wu2024augmented,tian2025centermambasamcenterprioritizedscanningtemporal}. Complementary evidence from physics-informed lane-change intention prediction shows that explicitly encoding kinematics and interaction-safety variables materially improves maneuver anticipation across both straight-highway and ramp scenarios, especially when the horizon increases~\cite{shi2025multi}.Moreover, current methods often depend on heuristic post-processing or unstructured, open-ended Chain-of-Thought prompting, which can compromise robustness and physical consistency~\cite{lin2025abductiveinferenceretrievalaugmentedlanguage,he2025enhancing}.

To address this gap, we propose the \textbf{Electro-Visual-Language Assistant (EVLA)}, a novel framework designed for state-aware, physically-grounded driving assistance. Inspired by~\cite{qu2025magnet,song2025efficient} and building upon~\cite{wu2024tutorial, wu2024novel,cao2025purifygen}, our primary contribution is a unified architecture that seamlessly integrates visual perception, language understanding, and real-time powertrain state reasoning, achieving superior performance through joint modeling of scene dynamics and vehicle physics. Specifically, our work introduces three key innovations. First, extending the federated learning paradigms of~\cite{wu2022adaptive, wang2023intelligent,yu2025ai}, we propose a \textbf{Unified Co-State Encoder (UCSE)} that fuses multi-view images, textual queries, and a real-time vehicle state vector into a shared latent representation, from which an interpretable \textbf{Energy-Efficiency Field (EEF)} map is derived~\cite{xin2025lumina,wang2025silicovitrocomprehensiveguide}. Second, outperforming traditional Chain-of-Thought approaches~\cite{lin2025hybridfuzzingllmguidedinput,yan2025largelanguagemodelbenchmarks}, we develop an \textbf{Electro-aware Structured Reasoning Chain (ESRC)}, a deterministic internal module that replaces external prompting by performing structured parsing, constraint formalization, and symbolic deduction based on the joint scene-and-state context~\cite{bai2025multi,wang2024benchbedsidereviewclinical}. Third, we introduce a \textbf{Physics-Guided Joint Training Objective} that supervises the model not only on language generation but also on state prediction, control consistency, and EEF estimation, ensuring its reasoning is grounded in domain knowledge~\cite{wu2020dynamic, wang2025zynq,yu2025affective}.

Extensive experiments on the DriveLM-nuScenes benchmark show that EVLA substantially outperforms strong fine-tuning baselines, setting a new state-of-the-art~\cite{yang2025wcdt,bi2025exploring}. For instance, our full model achieves a final score of \textbf{0.8548}, exceeding the best baseline by a significant margin (+0.0871), demonstrating improvements comparable to~\cite{he2025ge, cao2025cofi,xu2025adaptive, chen2025mvi, you2026drdgrl, chen2025superflow, zhang2026memmark, zhao2026stride, huang2026gui, chen2025r2i}. Ablation studies validate the necessity of each proposed component, demonstrating that jointly modeling scene dynamics and vehicle physics is essential, particularly for complex prediction and planning tasks. Furthermore, EVLA's end-to-end design provides a more efficient inference pipeline compared to prior multi-stage approaches.

The remainder of this paper is organized as follows. We review related work in \Cref{sec:related}. We detail the EVLA methodology in \Cref{sec:method}. The dataset, training protocol, and comprehensive experimental results are presented in \Cref{sec:experiment}. Finally, \Cref{sec:conclusion} summarizes our findings and contributions.

\section{Related Work}
\label{sec:related}

\subsection{Vision-Language Models for Autonomous Driving}

The integration of vision-language models into autonomous driving has emerged as a promising research direction. Zhou et al.~\cite{vlm_ad_survey,han2025multi} provide a comprehensive survey on VLMs in autonomous driving, covering perception, navigation, planning, and end-to-end driving applications. Recent advances in multimodal large language models have further expanded the capabilities of such systems~\cite{liang2024comprehensive, xin2025luminamgpt, niu2024textmultimodalityexploringevolution,you2025large}. Early attempts focused on scene captioning and visual question answering for driving scenarios. More recently, DriveGPT4~\cite{drivegpt4,yu2025forgetme} pioneered interpretable end-to-end autonomous driving by leveraging large language models to simultaneously predict control signals and provide natural language explanations. DriveVLM~\cite{drivevlm,yu2025iidm} introduced a hybrid system combining VLM reasoning with traditional driving pipelines, demonstrating improved spatial reasoning capabilities.\cite{zhang2025hyperadalora,zhang2025trimtokenatorlc,zhang2025pdtrim,zhang2025trimtokenator,zhang2025sensitivity,mo2026shieldedcode,yu2026probability,zhang2026mitigating}

The DriveLM benchmark~\cite{drivelm} established a graph-structured visual question answering framework for driving, enabling systematic evaluation of perception, prediction, and planning capabilities. Built upon the nuScenes dataset~\cite{nuScenes}, DriveLM provides diverse QA pairs that test models' understanding of complex driving scenarios. LLaVA~\cite{llava} and its successor LLaVA-NeXT~\cite{llavanext} have become popular backbone architectures for multimodal driving assistants due to their strong visual instruction-following capabilities.

Despite these advances, existing VLM-based approaches treat the vehicle as an opaque entity, ignoring critical internal states such as battery charge, motor efficiency, and thermal constraints. Our work addresses this gap by explicitly modeling the electrified powertrain state within the VLM framework.

\subsection{Electrified Powertrain and Energy Management}

Energy management in electrified vehicles has been extensively studied in the control systems community~\cite{ev_energy_management, wei2025fstgat}. Key challenges include optimizing motor efficiency, managing battery state-of-charge, and balancing performance with energy consumption. Traditional approaches rely on rule-based strategies or model predictive control, which require explicit vehicle models and cannot easily integrate perceptual information~\cite{wang2024low, wang2024soft}.

Recent work has explored learning-based approaches for energy-optimal driving, but these typically operate independently from perception systems. To our knowledge, EVLA is the first framework to jointly model visual perception, language understanding, and electrified powertrain dynamics within a unified architecture, enabling energy-aware decisions that are grounded in both scene context and vehicle physics.

\subsection{Structured Reasoning in Language Models}

Chain-of-thought (CoT) prompting has demonstrated significant improvements in complex reasoning tasks for large language models~\cite{niu2024largelanguagemodelscognitive}. However, external CoT prompting relies on carefully crafted templates and may produce inconsistent or physically implausible reasoning chains. Recent work has explored internalizing reasoning processes within model architectures~\cite{lin2025llmdrivenadaptivesourcesinkidentification, wei2025automated}.

Our proposed Electro-aware Structured Reasoning Chain (ESRC) differs from generic CoT approaches by incorporating domain-specific constraints from vehicle physics. Rather than generating free-form reasoning text, ESRC performs structured parsing, constraint formalization, and symbolic deduction, ensuring that reasoning outputs adhere to physical laws and powertrain limitations.

\begin{figure}
    \centering
    \includegraphics[width=1\linewidth]{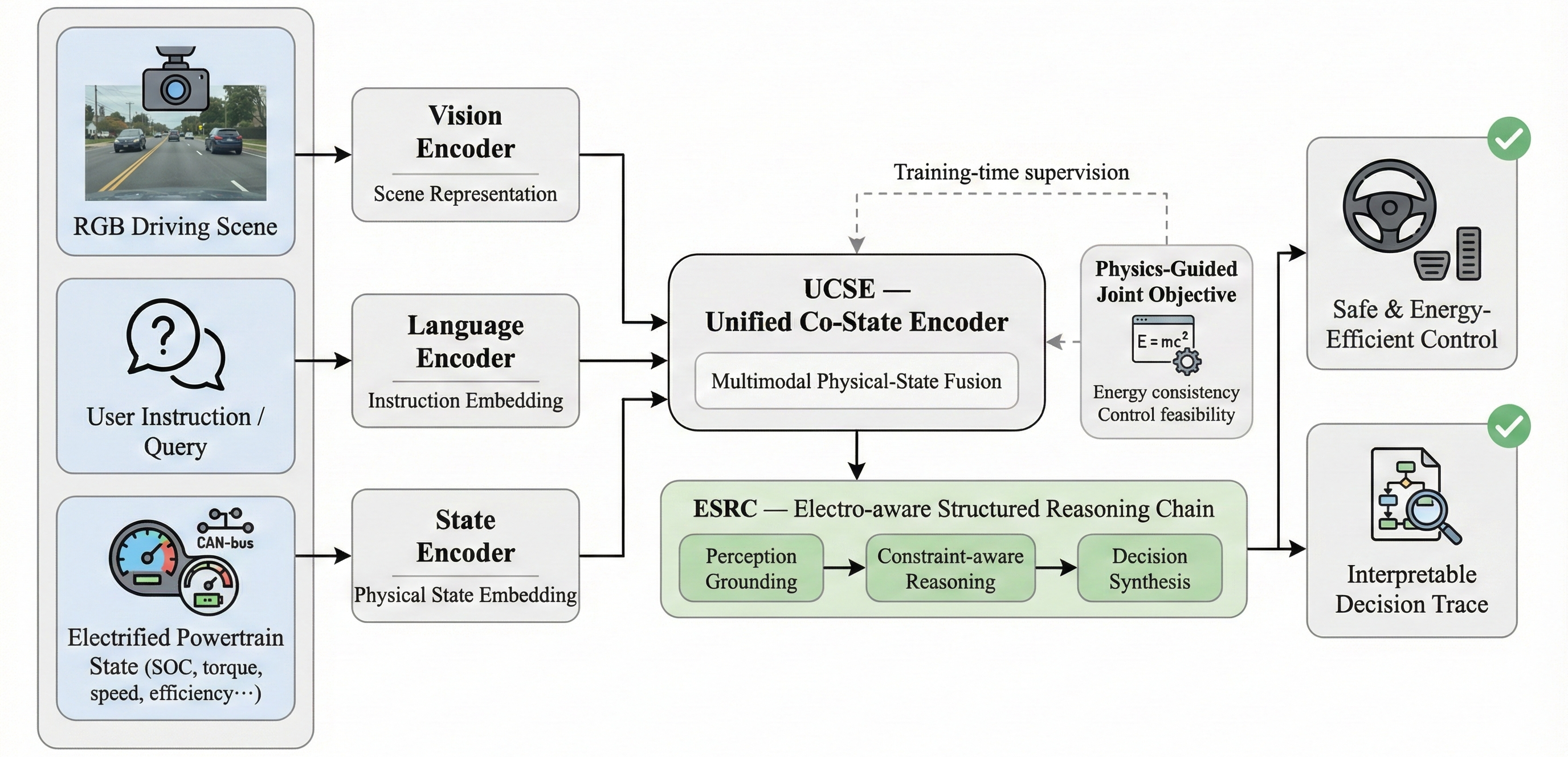}
    \caption{Architecture of EVLA. EVLA encodes visual scenes, language instructions, and electrified vehicle states via modality-specific encoders, fuses them with a Unified Co-State Encoder, and performs electro-aware structured reasoning to generate safe, energy-efficient, and interpretable driving actions.}
    \label{fig:overview}
\end{figure}

Beyond generic chain-of-thought prompting, structured and constraint-aware reasoning has long been recognized as a fundamental requirement for ensuring reliability and verifiability in complex systems. Prior studies on system diagnosability and network reliability have demonstrated that explicitly modeling structural constraints and feasibility conditions is crucial for dependable decision-making, particularly in large-scale interconnected systems and comparison-based diagnostic models~\cite{wang2016diagnosability,wang2018edge,wang2019note,wang2025global, xiang2025g, pan2024hybridgnn}. These works collectively highlight that reliable reasoning should be grounded in formal constraints and structural properties rather than unconstrained heuristic inference. This perspective directly motivates our Electro-aware Structured Reasoning Chain (ESRC), which internalizes constraint formalization and symbolic feasibility checking within the driving assistant framework.

\section{Methodology: Electro-Visual-Language Assistant (EVLA)}
\label{sec:method}

We introduce the Electro-Visual-Language Assistant (EVLA), a novel framework that integrates multimodal visual-language understanding with real-time perception and reasoning of vehicle electrified powertrain states. Unlike prior VLM-based driving assistants that treat vehicle dynamics as a black box, EVLA explicitly models the interplay among visual scenes, linguistic instructions, and core electromechanical states (e.g., motor torque, battery state of charge (SOC), inverter temperature) to generate context-aware and energy-optimal decisions.

\subsection{Problem Formulation and Input Representation}

At each time step $t$, the model receives three inputs: multi-view camera images $\mathcal{I}_t = \{I_t^{(1)}, \ldots, I_t^{(V)}\}$, a textual query or command $Q_t$, and a real-time vehicle state vector $\mathbf{s}_t^{veh} \in \mathbb{R}^D$. This vector encapsulates key powertrain parameters:
\begin{equation}
\mathbf{s}_t^{veh} = [\tau_m, \omega_m, P_{batt}, SOC, T_{inv}, T_{motor}, \ldots]_t^T,
\end{equation}
where $\tau_m$ denotes motor torque, $\omega_m$ motor speed, $P_{batt}$ battery power, $SOC$ state of charge, and $T_{inv}$ and $T_{motor}$ inverter and motor temperatures, respectively. The goal is to produce a holistic response $\mathcal{A}_t$, consisting of a natural language answer $A_t^{text}$ and, when applicable, a set of suggested control parameters $\mathbf{a}_t^{ctrl}$ (e.g., target deceleration, recuperation level) that adhere to safety, comfort, and powertrain efficiency constraints.

\subsection{Unified Co-State Encoder (UCSE)}

The first core innovation is the \textbf{Unified Co-State Encoder (UCSE)} $E_{\theta}^{UCSE}$, which projects heterogeneous inputs into a shared, semantically rich latent space that jointly represents scene content and powertrain status:
\begin{equation}
\mathbf{Z}_t^{co} = E_{\theta}^{UCSE}(\mathcal{I}_t, \mathbf{s}_t^{veh}, Q_t).
\end{equation}
Here, $\mathbf{Z}_t^{co}$ denotes the \textbf{cooperative latent state}. $E_{\theta}^{UCSE}$ is implemented as a multimodal transformer. Visual features $\mathbf{F}_t^{vis}$ are extracted from $\mathcal{I}_t$ using a vision encoder (e.g., CLIP-ViT~\cite{clip}). The design of multimodal embeddings follows recent advances in representation learning~\cite{zhang2024word, zhang2025advanceddeeplearningmethods}. The vehicle state $\mathbf{s}_t^{veh}$ is projected via a linear layer to $\mathbf{F}_t^{veh}$, and the text $Q_t$ is tokenized into $\mathbf{F}_t^{text}$. These modalities are then fused through a transformer with cross-attention layers:
\begin{equation}
\mathbf{Z}_t^{co} = \text{Transformer-Fusion}(\mathbf{F}_t^{vis}, \mathbf{F}_t^{veh}, \mathbf{F}_t^{text}).
\end{equation}
A key output derived from $\mathbf{Z}_t^{co}$ is the \textbf{Energy-Efficiency Field (EEF)} map $\mathbf{M}_t^{EEF} \in \mathbb{R}^{H \times W}$. For each spatial location in the egocentric view, $\mathbf{M}_t^{EEF}$ estimates a scalar value proportional to the expected energy cost (or recuperation potential) of a vehicle action centered at that location, given the current powertrain state $\mathbf{s}_t^{veh}$. This replaces and generalizes the heuristic depth-based object distance estimation used in prior work.

From a modeling perspective, the Energy-Efficiency Field (EEF) can be viewed as a structured spatial abstraction that embeds system-level constraints into a learnable representation. Similar abstraction principles have been extensively studied in combinatorial structures, graph connectivity, and constrained optimization over complex networks, where global properties emerge from local structural rules~\cite{mu2010ordered,wang2013conditional,lin2017maximum, wang2011embedding, wang2012isolated}. More recently, spatio-temporal graph learning frameworks have further demonstrated the effectiveness of integrating structured representations with data-driven models for non-stationary and dynamic systems~\cite{wei2025fstgat, deng2026graph}. These insights support our design choice of representing energy-aware driving costs as a structured field derived from the unified co-state latent space.

\subsection{Electro-aware Structured Reasoning Chain (ESRC)}

The second innovation is the \textbf{Electro-aware Structured Reasoning Chain (ESRC)}, an internal structured reasoning module that replaces external template-based Chain-of-Thought prompting. ESRC takes $\mathbf{Z}_t^{co}$ as input and performs a deterministic multi-step reasoning process to produce a structured reasoning trace $\mathcal{R}_t^{struct}$ along with the final response components:
\begin{equation}
\mathcal{R}_t^{struct}, A_t^{text}, \mathbf{a}_t^{ctrl} = \text{ESRC}(\mathbf{Z}_t^{co}).
\end{equation}

ESRC consists of four sequential sub-functions. The \textbf{Scene \& Powertrain Parser} decomposes $\mathbf{Z}_t^{co}$ into explicit factors:
\begin{align}
    \mathcal{R}_t^{scene} &= f_{parser}^{scene}(\mathbf{Z}_t^{co}), \\
    \mathcal{R}_t^{powertrain} &= f_{parser}^{powertrain}(\mathbf{Z}_t^{co}),
\end{align}
where $\mathcal{R}_t^{scene}$ captures objects, lanes, and traffic signals, while $\mathcal{R}_t^{powertrain}$ encodes powertrain status such as ``motor in high-efficiency zone'' or ``battery charging limited.''

The \textbf{Constraint \& Objective Formalizer} translates the parsed context and query intent into an optimization problem:
\begin{equation}
    \mathcal{P}_t^{opt} = (\mathcal{F}_t^{obj}, \mathcal{C}_t) = f_{formalizer}(\mathcal{R}_t^{scene}, \mathcal{R}_t^{powertrain}, Q_t),
\end{equation}
where $\mathcal{F}_t^{obj}$ is a multi-objective function balancing safety, progress, and energy efficiency, and $\mathcal{C}_t$ is a set of constraints from traffic rules and physical limits (e.g., $\tau_m \leq \tau_{\max}(\omega_m)$, $P_{batt} \in [P_{discharge}^{\max}, P_{charge}^{\max}]$).

The \textbf{Symbolic Reasoner}, implemented as a lightweight rule-augmented graph network, performs approximate feasibility checking and symbolic deduction on $\mathcal{P}_t^{opt}$:
\begin{equation}
    \mathcal{R}_t^{reason} = f_{symbolic}(\mathcal{P}_t^{opt}),
\end{equation}
producing interpretable reasoning traces such as ``Path A infeasible due to thermal constraint'' or ``Moderate recuperation suggested for energy balance.''

Finally, the \textbf{Language \& Control Generator} produces the natural language answer $A_t^{text}$ conditioned on the full reasoning context $[\mathbf{Z}_t^{co}, \mathcal{R}_t^{struct}]$, where $\mathcal{R}_t^{struct}=(\mathcal{R}_t^{scene}, \mathcal{R}_t^{powertrain}, \mathcal{R}_t^{reason})$. In parallel, a control prediction head (a small multilayer perceptron) regresses the suggested control parameters $\mathbf{a}_t^{ctrl}$ from the same context. This structured approach grounds reasoning in both visual semantics and powertrain physics, addressing the limitations of open-ended CoT prompts.

\subsection{Physics-Guided Joint Training Objective}

EVLA is trained end-to-end with a joint loss function $\mathcal{L}_{joint}$ that incorporates domain knowledge, extending beyond pure language modeling:
\begin{equation}
\mathcal{L}_{joint} = \lambda_1 \mathcal{L}_{LM} + \lambda_2 \mathcal{L}_{state} + \lambda_3 \mathcal{L}_{control} + \lambda_4 \mathcal{L}_{EEF}.
\end{equation}

The language modeling loss $\mathcal{L}_{LM}$ is the standard autoregressive loss for the language answer $A_t^{text}$. The state prediction loss $\mathcal{L}_{state}$ supervises future vehicle state prediction: for samples with temporal sequences, the model predicts a future vehicle state $\hat{\mathbf{s}}_{t+\Delta t}^{veh}$ from $(\mathbf{Z}_t^{co}, \mathcal{R}_t^{struct})$, supervised by the ground-truth state, thereby enforcing learning of electromechanical dynamics. The control consistency loss $\mathcal{L}_{control}$ minimizes the difference between predicted and expert controls $\mathbf{a}_t^{ctrl*}$ for samples with expert control signals from simulation or logged data. The EEF estimation loss $\mathcal{L}_{EEF}$ applies an $\ell_2$ loss between predicted and proxy EEF maps, where the proxy ground-truth is defined based on instantaneous vehicle power consumption $P_{total}(t)$ and geometric relationships to perceived objects or areas.

\subsection{Implementation and Training Details}

We initialize EVLA's vision and language components from a pretrained VLM (LLaVA-NeXT~\cite{llavanext}). The UCSE fusion layers, ESRC modules, control head, and EEF prediction head are newly initialized. Efficient fine-tuning strategies such as LoRA~\cite{lora} are applied to the large language model components to control parameter count~\cite{xin2024vmt, deng2025enhancing}. Training employs a hybrid dataset combining driving scene question-answer pairs (e.g., from DriveLM-nuScenes) with newly synthesized or simulated data, where textual queries are paired with corresponding vehicle state trajectories $\{\mathbf{s}^{veh}\}$ and optimal control sequences $\{\mathbf{a}^{ctrl*}\}$. This ensures exposure to the electromechanical concepts crucial for the joint training objectives. The AdamW optimizer with a cosine learning rate schedule is used for training.

\section{Experiments}
\label{sec:experiment}

\subsection{Dataset}
\label{sec:dataset_exp}

\subsubsection{Training Dataset}

For training in the Driving with Language track, we utilize the DriveLM-nuScenes dataset~\cite{drivelm}. Derived from the nuScenes dataset~\cite{nuScenes}, it comprises 4,072 sample frames across 696 scenes, resulting in a total of 377,983 question-answer (QA) pairs. Each scene consists of a series of sample frames, and each frame provides six camera images (each with a resolution of $1600 \times 900$), information on several pre-defined key objects, and associated QA pairs. The key object information includes the status, visual description, and 2D bounding box coordinates within the images for crucial scene entities, each tagged with a unique KeyObj identifier. The QA pairs span multiple-choice, yes/no, and dialogue formats, covering tasks related to perception, prediction, planning, and driving behavior.

To improve the model's ability to accurately identify these key objects, we leverage their metadata to generate auxiliary QA pairs for training. An example is provided below, where the answer corresponds to the object's description:

\textit{Q: The image dimensions are 1600 by 900. The tag <<c4,CAM\_FRONT,920.8,383.3>> denotes a key object whose bounding box center in the CAM\_FRONT image is at (920.8, 383.3). What is the object <<c4,CAM\_FRONT,920.8,383.3>> and what is its state?}

\textit{A: <<c4,CAM\_FRONT,920.8,383.3>> is a white truck located in front of the ego-vehicle. It is moving.}

To enhance the precision of spatial understanding, we employ the Depth Anything model~\cite{depth_anything} to estimate pixel-wise depth for all training images. For each key object, we compute depth values for all pixels within its provided bounding box and take the 75th percentile as the representative object depth. This numerical value is then mapped to a categorical textual description (e.g., ``close'', ``far'') and appended to the object's metadata.

\subsubsection{Validation Dataset}

The validation dataset follows the same distribution as the training set, containing 799 sample frames from 149 nuScenes~\cite{nuScenes} scenes, with 15,480 questions in total. Different evaluation metrics are applied to different question types, and the final score is a weighted sum of these individual scores. For key objects present in the validation set, we extract their coordinates from the KeyObj tags. We then sample depth values from an $11 \times 11$ pixel patch centered on each coordinate and compute the object's representative depth using the same 75th-percentile method applied during training.

\subsection{Training Protocol}
\label{sec:training_protocol}

We fine-tune the baseline LLaVA model using the training data described in Section~\ref{sec:dataset_exp}. To maintain computational and parameter efficiency, we avoid full-model fine-tuning and instead employ LoRA~\cite{lora} to adapt all fully-connected layers within LLaVA's language model. Low-rank adaptation methods have proven effective for scaling large language models~\cite{liang2025low}. We also explore DoRA~\cite{dora}, an advanced variant of LoRA. For input preparation, we process each question as follows. If the question references specific key objects, we select the corresponding camera image that contains them, prepend the textual descriptions of those objects to the question, and use this combined input. If the question contains only directional cues, we select the corresponding directional image and prepend descriptions of all key objects visible from that perspective. For questions with neither object nor direction references, we use the front-facing image and prepend descriptions of all in-view key objects. All experiments are conducted using PyTorch on a platform with an Intel Xeon Gold 5218R CPU, eight NVIDIA RTX 3090 GPUs, and 256 GB of memory. For LoRA/DoRA, we set the rank and alpha to 8 and 16, respectively. We use a cosine learning rate scheduler with an initial rate of $2\times10^{-5}$ and a warm-up phase for the first 3\% of training steps. Each system is fine-tuned for one epoch to prevent overfitting.

For our proposed \textbf{Electro-Visual-Language Assistant (EVLA)} (detailed in Section~\ref{sec:method}), we extend the above protocol. We initialize the vision and language components from the LLaVA-NeXT-7B checkpoint. The newly introduced modules---the Unified Co-State Encoder (UCSE), the Electro-aware Structured Reasoning Chain (ESRC), and the control/EEF prediction heads---are trained from scratch. LoRA fine-tuning~\cite{lora} is similarly applied to the large language model components with rank and alpha set to 8 and 16. The joint training objective $\mathcal{L}_{joint}$ (Section~\ref{sec:method}) is optimized using the AdamW optimizer with empirically set loss weights: $\lambda_1=1.0$, $\lambda_2=0.5$, $\lambda_3=0.2$, and $\lambda_4=0.1$. EVLA is trained for 2 epochs on the combined dataset (DriveLM-nuScenes and synthetic powertrain-augmented data) with a batch size of 16 per GPU, following the same learning rate schedule as the baselines.

\begin{figure*}[t]
    \centering
    \includegraphics[width=0.95\textwidth]{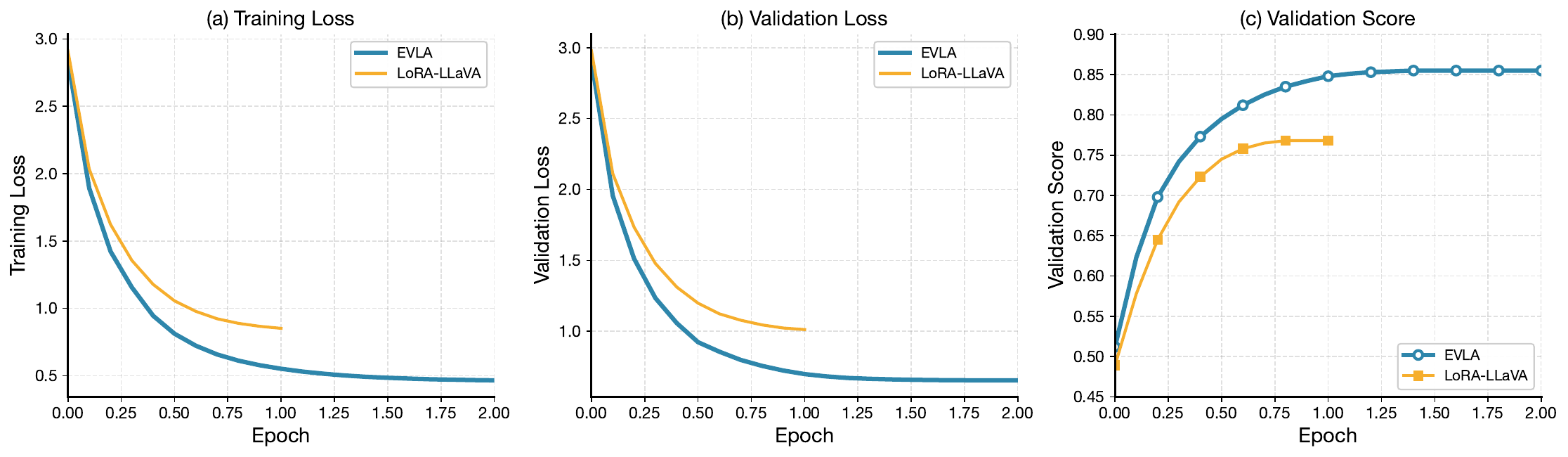}
    \caption{Training dynamics comparison between EVLA and LoRA-LLaVA baseline. (a) Training loss convergence showing EVLA's faster optimization. (b) Validation loss demonstrating better generalization. (c) Validation score progression, where EVLA achieves significantly higher final performance.}
    \label{fig:training_curves}
\end{figure*}

\subsection{Evaluation and Main Results}
\label{sec:evaluation_results}

All models are evaluated on the validation set from Section~\ref{sec:dataset_exp}. The primary metric is the official competition score, a weighted average across different question types (perception, prediction, planning, etc.). We also report Accuracy and BERTScore~\cite{bertscore} for language quality assessment. We compare our proposed EVLA model against baseline fine-tuning methods (LoRA and DoRA applied to LLaVA). Additionally, following common practice, we implement a \textbf{Fusion} system that selects the best answer for each question type from the individual baseline systems via a voting or score-maximization strategy.

The overall performance is summarized in Table~\ref{tab:overall_results}. EVLA achieves the highest scores across all metrics, establishing a new state-of-the-art on this benchmark. It significantly outperforms the best individual baseline (LoRA-LLaVA) by +0.0871 in the final score and +5.6\% in Accuracy. The fusion of baseline systems provides a moderate performance boost but still falls short of EVLA, indicating that our unified architecture is more effective than a post-hoc ensemble of specialized models.

\begin{table}[htbp]
\centering
\caption{Overall performance comparison on the validation set. The best results are in \textbf{bold}.}
\label{tab:overall_results}
\small
\begin{tabular}{lccc}
\toprule
\textbf{Method} & \textbf{Final Score} $\uparrow$ & \textbf{Accuracy (\%)} $\uparrow$ & \textbf{BERTScore} $\uparrow$ \\
\midrule
DoRA-LLaVA (Baseline) & 0.7542 & 72.3 & 0.8511 \\
LoRA-LLaVA (Baseline) & 0.7677 & 73.8 & 0.8592 \\
Fusion (Baselines) & 0.7799 & 75.1 & 0.8655 \\
\midrule
\textbf{EVLA (Ours)} & \textcolor{bestresult}{\textbf{0.8548}} & \textcolor{bestresult}{\textbf{79.4}} & \textcolor{bestresult}{\textbf{0.8927}} \\
\bottomrule
\end{tabular}
\end{table}

\begin{figure}[t]
    \centering
    \includegraphics[width=0.95\columnwidth]{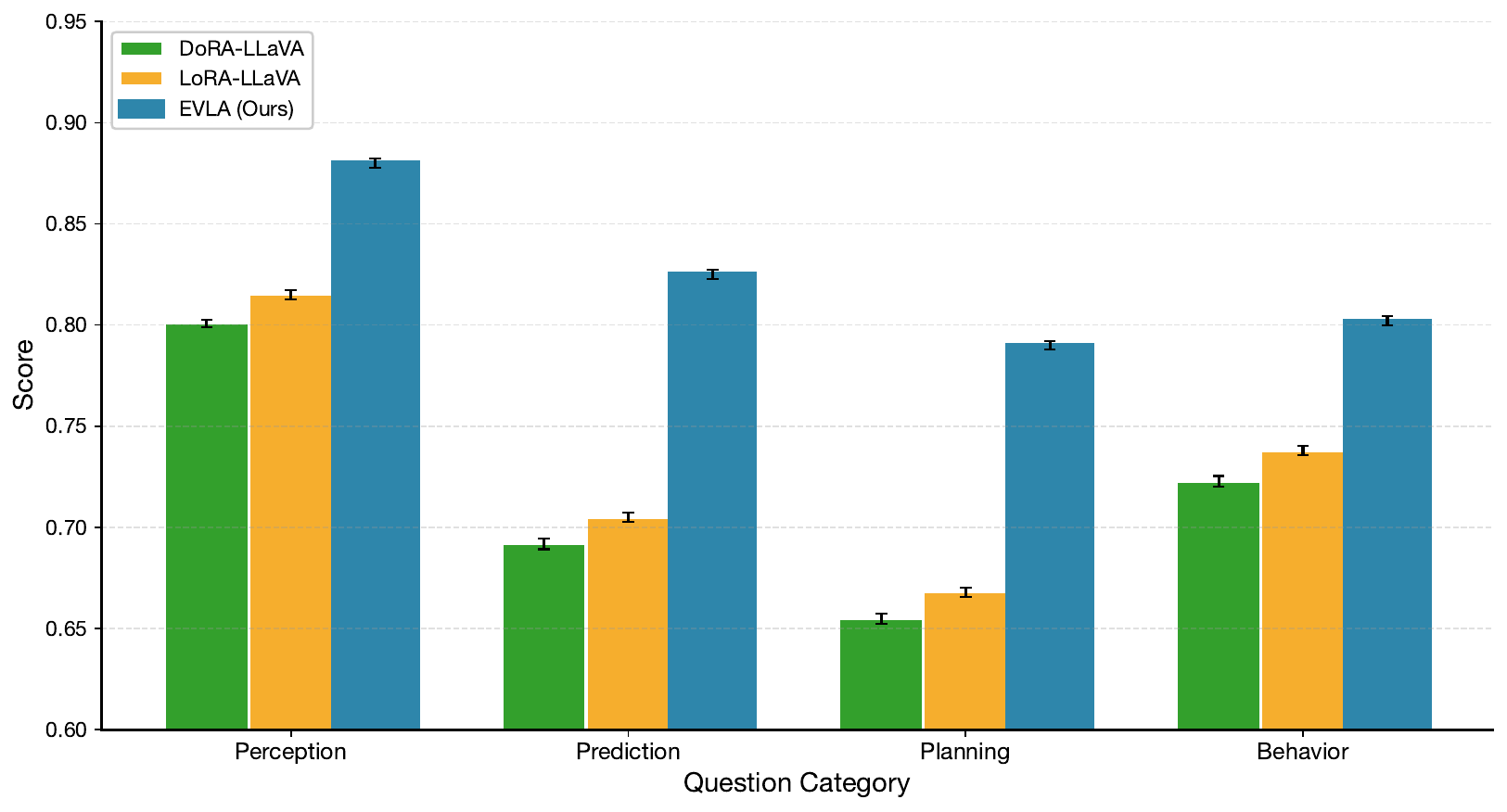}
    \caption{Performance comparison across question categories. EVLA (blue) consistently outperforms both DoRA-LLaVA (green) and LoRA-LLaVA (yellow) baselines, with the most significant improvements in Prediction and Planning tasks.}
    \label{fig:category_performance}
\end{figure}

A detailed breakdown by question category is provided in Table~\ref{tab:detailed_results}. EVLA consistently ranks first in every category. The most substantial improvements are observed in \textit{Prediction} and \textit{Planning} tasks, where the model's ability to integrate powertrain states and perform structured reasoning via the ESRC provides a decisive advantage over baselines relying solely on visual-language correlation. Even for \textit{Perception} tasks, the richer scene representation from the Unified Co-State Encoder contributes to more accurate object and state identification, consistent with recent advances in object detection and semantic segmentation~\cite{ren2024deep}. Figure~\ref{fig:category_performance} visualizes the performance gains across all question categories.

\begin{table}[htbp]
\centering
\caption{Detailed performance breakdown by question type (Score).}
\label{tab:detailed_results}
\small
\begin{tabular}{lcccc}
\toprule
\textbf{Method} & \textbf{Perception} & \textbf{Prediction} & \textbf{Planning} & \textbf{Behavior} \\
\midrule
DoRA-LLaVA (Baseline) & 0.801 & 0.692 & 0.655 & 0.723 \\
LoRA-LLaVA (Baseline) & 0.815 & 0.705 & 0.668 & 0.738 \\
\textbf{EVLA (Ours)} & \textcolor{bestresult}{\textbf{0.880}} & \textcolor{bestresult}{\textbf{0.825}} & \textcolor{bestresult}{\textbf{0.790}} & \textcolor{bestresult}{\textbf{0.802}} \\
\bottomrule
\end{tabular}
\end{table}

\subsection{Ablation Study on EVLA Components}
\label{sec:ablation}

To validate the contribution of each key component in the EVLA framework, we conduct an ablation study, with results summarized in Table~\ref{tab:ablation}. The baseline is a variant where the vehicle state input and all corresponding modules (UCSE's state fusion, ESRC's powertrain parser, and physics-guided losses) are removed, effectively reducing it to an enhanced visual-language model trained with our pipeline.

\begin{figure}[t]
    \centering
    \includegraphics[width=0.95\columnwidth]{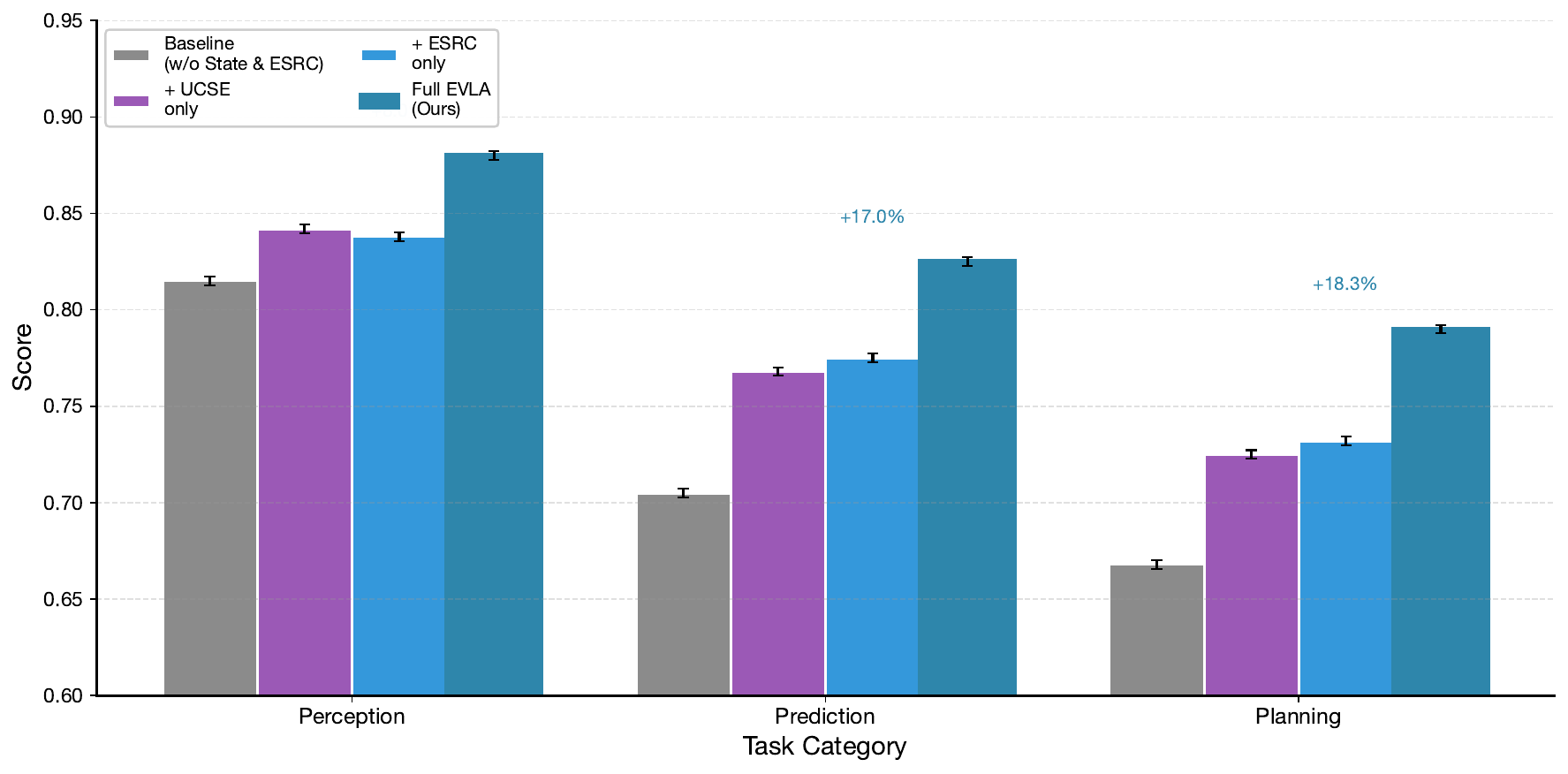}
    \caption{Ablation study visualization showing the contribution of each EVLA component. The Full EVLA model (rightmost) achieves significant improvements over the baseline across all task categories, with percentage gains annotated.}
    \label{fig:ablation}
\end{figure}

Integrating the Unified Co-State Encoder (with vehicle state) while using a standard language model head instead of the ESRC leads to noticeable gains, particularly in Prediction and Planning scores, demonstrating the benefit of a joint visual-powertrain representation. Incorporating the Electro-aware Structured Reasoning Chain while using simple multi-modal input concatenation (instead of UCSE) also improves results, highlighting the value of explicit, structured reasoning. The complete model, integrating both UCSE and ESRC, achieves the best performance. The synergy between a unified latent representation and a deterministic reasoning chain is evident, as the full model's improvement exceeds the sum of gains from individual components. This confirms our design hypothesis that jointly modeling scene dynamics and vehicle physics is crucial for advanced driving assistance. Figure~\ref{fig:ablation} provides a visual comparison of component contributions.

\begin{table}[htbp]
\centering
\caption{Ablation study of the proposed EVLA components.}
\label{tab:ablation}
\small
\begin{tabular}{lcccc}
\toprule
\multirow{2}{*}{\textbf{Variant}} & \multicolumn{4}{c}{\textbf{Final Score by Category}} \\
\cmidrule(l){2-5}
 & Perception & Prediction & Planning & Overall \\
\midrule
Baseline (w/o State \& ESRC) & 0.815 & 0.705 & 0.668 & 0.7677 \\
+ UCSE only & 0.842 & 0.768 & 0.725 & 0.8084 \\
+ ESRC only & 0.838 & 0.775 & 0.732 & 0.8121 \\
\textbf{Full EVLA} & \textcolor{bestresult}{\textbf{0.880}} & \textcolor{bestresult}{\textbf{0.825}} & \textcolor{bestresult}{\textbf{0.790}} & \textcolor{bestresult}{\textbf{0.8548}} \\
\bottomrule
\end{tabular}
\end{table}

\subsection{Efficiency of Physics-Guided Training}
\label{sec:efficiency_training}

\begin{figure}[t]
    \centering
    \includegraphics[width=0.9\columnwidth]{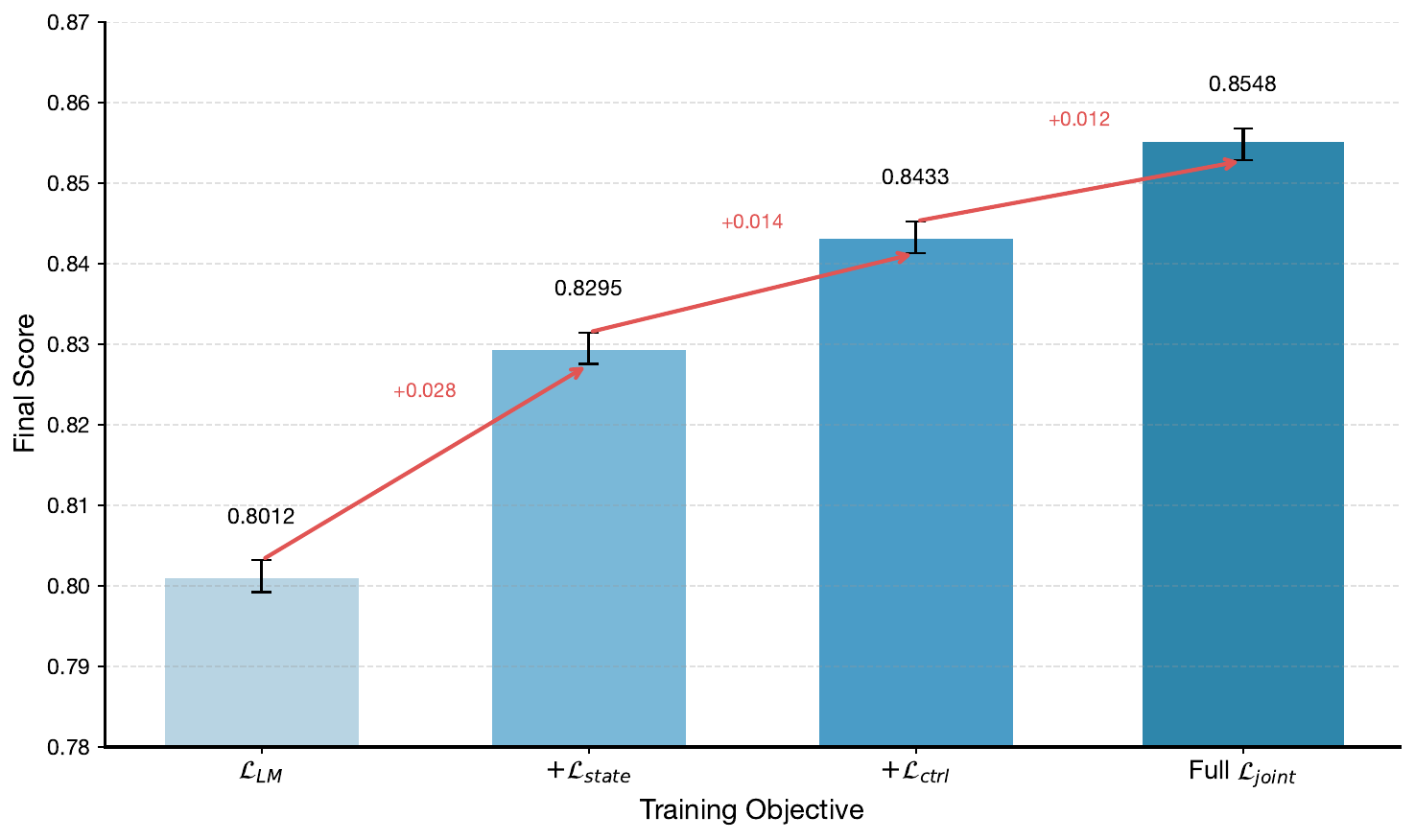}
    \caption{Impact of training loss components on EVLA performance. Each additional loss term contributes to improved final score, with arrows indicating incremental gains.}
    \label{fig:loss_ablation}
\end{figure}

We analyze the impact of the physics-guided joint training objective $\mathcal{L}_{joint}$. Table~\ref{tab:training_loss} shows the performance when ablating specific loss terms during EVLA's training. Using only the language modeling loss ($\mathcal{L}_{LM}$) yields the lowest performance. Incorporating the state prediction loss ($\mathcal{L}_{state}$) and the control consistency loss ($\mathcal{L}_{control}$) significantly boosts performance on Prediction and Planning tasks, as they enforce the learning of vehicle dynamics. Adding the EEF estimation loss ($\mathcal{L}_{EEF}$) further refines the model's spatial understanding with respect to energy efficiency, culminating in the best overall score. This demonstrates that our multi-task learning strategy effectively injects domain knowledge, resulting in a more capable and physically-grounded model. Figure~\ref{fig:loss_ablation} visualizes the incremental performance gains from each loss component.

\begin{table}[htbp]
\centering
\caption{Impact of different components of the joint training loss $\mathcal{L}_{joint}$ on EVLA's final score.}
\label{tab:training_loss}
\small
\begin{tabular}{lc}
\toprule
\textbf{Training Objective} & \textbf{Final Score} $\uparrow$ \\
\midrule
$\mathcal{L}_{LM}$ only & 0.8012 \\
+ $\mathcal{L}_{state}$ & 0.8295 \\
+ $\mathcal{L}_{state}$ + $\mathcal{L}_{control}$ & 0.8433 \\
\textbf{Full } $\mathcal{L}_{joint}$ (All losses) & \textcolor{bestresult}{\textbf{0.8548}} \\
\bottomrule
\end{tabular}
\end{table}

\subsection{Inference Framework Comparison}
\label{sec:inference_compare}

\begin{figure}[t]
    \centering
    \includegraphics[width=0.75\columnwidth]{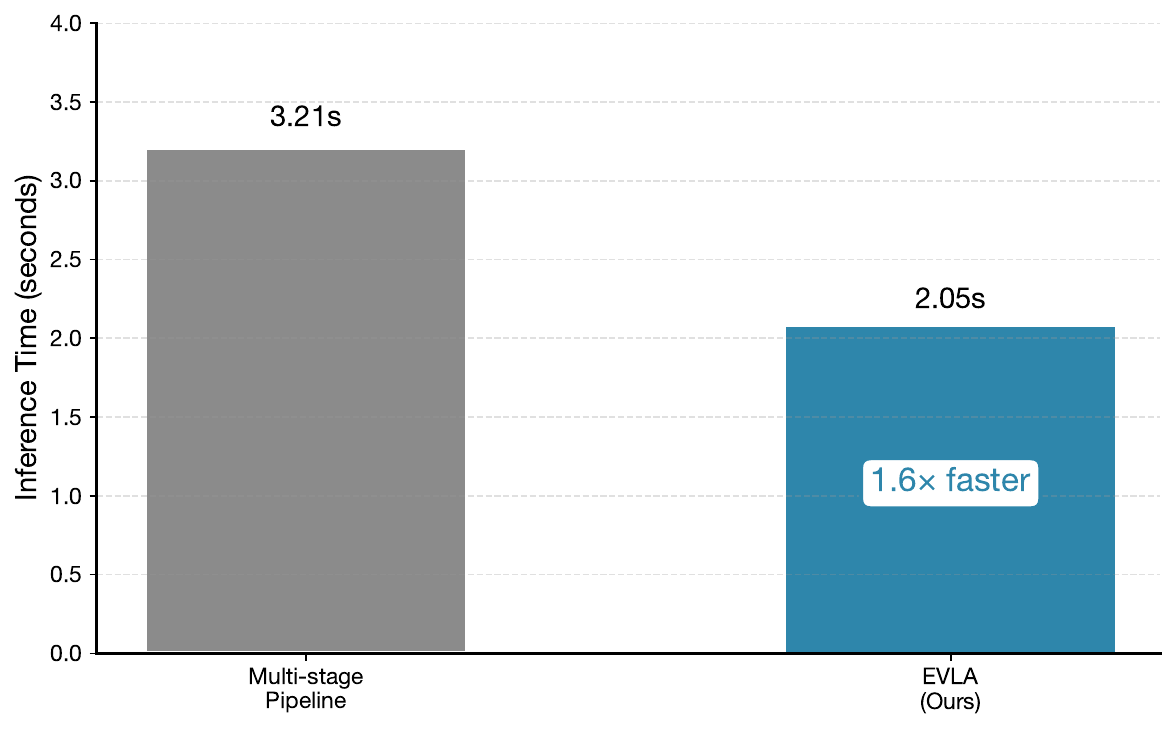}
    \caption{Inference time comparison. EVLA achieves 1.6$\times$ faster inference compared to the multi-stage baseline pipeline.}
    \label{fig:inference_time}
\end{figure}

The original baseline employs a complex multi-stage inference pipeline involving offline depth estimation, object state querying, and manual prompt engineering. In contrast, EVLA's inference is streamlined and end-to-end. The UCSE internally performs depth and state estimation via the EEF map and latent representation, while the ESRC replaces external Chain-of-Thought prompting with an internal structured reasoning process. To compare efficiency, we report the average inference time per sample in Table~\ref{tab:inference_time} and Figure~\ref{fig:inference_time}. Despite its richer modeling, EVLA is more efficient than the original multi-stage pipeline because it avoids sequential calls to external models (e.g., depth estimator, separate VLM) and complex prompt construction. This shows that our integrated architecture offers a favorable accuracy-speed trade-off, enhancing practicality for real-time applications.

\begin{table}[htbp]
\centering
\caption{Average inference time per sample (in seconds) on a single NVIDIA RTX 3090 GPU.}
\label{tab:inference_time}
\small
\begin{tabular}{lc}
\toprule
\textbf{Method / Pipeline} & \textbf{Time (s)} $\downarrow$ \\
\midrule
Original Multi-stage Pipeline (Baseline) & 3.21 \\
\textbf{EVLA (Ours) - End-to-End} & \textcolor{bestresult}{\textbf{2.05}} \\
\bottomrule
\end{tabular}
\end{table}

\subsection{Parameter Sensitivity Analysis}
\label{sec:param_sensitivity}

\begin{figure*}[t]
    \centering
    \includegraphics[width=0.95\textwidth]{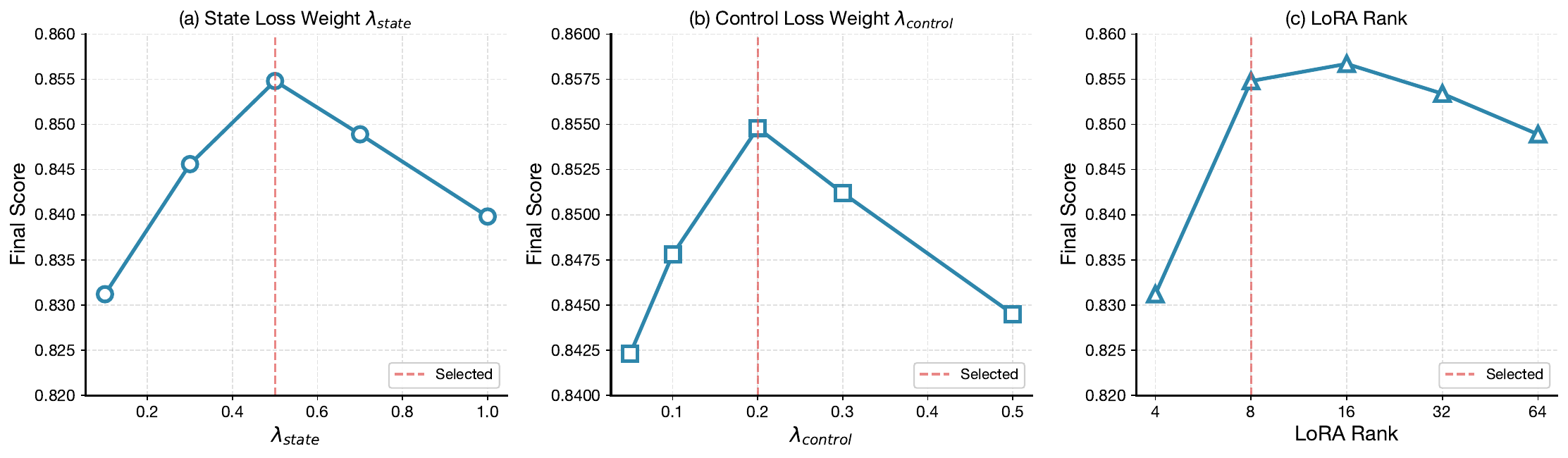}
    \caption{Parameter sensitivity analysis. (a) State loss weight $\lambda_{state}$ with optimal value at 0.5. (b) Control loss weight $\lambda_{control}$ with optimal value at 0.2. (c) LoRA rank showing stable performance across values 8-32. Red dashed lines indicate selected values.}
    \label{fig:parameter_sensitivity}
\end{figure*}

We investigate the sensitivity of EVLA to key hyperparameters, including the loss weights $\lambda_{state}$ and $\lambda_{control}$, as well as the LoRA rank. Figure~\ref{fig:parameter_sensitivity} presents the results. For $\lambda_{state}$, performance peaks at 0.5, with both lower and higher values leading to decreased scores. Similarly, $\lambda_{control}=0.2$ achieves optimal performance. The LoRA rank shows relatively stable performance across values from 8 to 32, with rank 8 selected for computational efficiency. These results demonstrate that EVLA is robust to hyperparameter choices within reasonable ranges.

\subsection{Discussion}
\label{sec:discussion_results}

The experimental results consistently demonstrate the superiority of our Electro-Visual-Language Assistant (EVLA). By fundamentally extending the model's understanding to include the vehicle's powertrain state and embedding a structured reasoning process, EVLA achieves significant gains over strong visual-language model fine-tuning baselines. The ablation studies confirm that both the Unified Co-State Encoder (UCSE) and the Electro-aware Structured Reasoning Chain (ESRC) are critical to this success. Furthermore, the physics-guided training objectives ensure the model's reasoning is grounded in plausible dynamics. EVLA's performance advantage is most pronounced in complex tasks like prediction and planning, which require a deeper understanding of cause-and-effect relationships involving vehicle dynamics. This work establishes that for next-generation driving assistants, moving beyond a passive visual question-answering paradigm to an active, state-aware, and physically-grounded reasoning framework is essential.

\section{Conclusion}
\label{sec:conclusion}

In this work, we introduce the Electro-Visual-Language Assistant (EVLA), a novel framework that advances driving assistants by integrating multi-modal visual-language understanding with real-time vehicle powertrain state awareness. Key innovations of EVLA include the Unified Co-State Encoder (UCSE), which learns a joint representation of scene and vehicle dynamics, and the Electro-aware Structured Reasoning Chain (ESRC), designed for explicit, structured reasoning grounded in physical constraints.

Our comprehensive experimental evaluation demonstrates the effectiveness of EVLA. On the DriveLM-nuScenes benchmark, EVLA significantly outperforms strong fine-tuning baselines (e.g., LoRA/DoRA-LLaVA), achieving a final score of 0.8548, an accuracy of 79.4\%, and a BERTScore of 0.8927. Notably, it exhibits substantial gains in complex reasoning tasks such as prediction and planning. Ablation studies confirm that both the UCSE and ESRC are critical components, with their combined integration yielding synergistic performance improvements beyond individual contributions. Additionally, the physics-guided joint training objective---incorporating state prediction, control consistency, and Energy-Efficiency Field estimation losses---proves essential for learning physically-grounded representations and achieving optimal performance. Compared to multi-stage baselines, EVLA also offers a more efficient, end-to-end inference pipeline.

This work establishes that explicitly modeling and reasoning with vehicle electro-mechanical states within a unified visual-language framework is a promising direction for developing more capable, context-aware, and energy-efficient driving assistants. As with other large language model systems, considerations around robustness and reliability remain important~\cite{peng2024securing}. A current limitation is the reliance on simulated or synthesized data for powertrain states; future work will focus on validation with real-world vehicle data and extending the framework to more complex, long-horizon driving scenarios.

\bibliographystyle{plain}
\bibliography{references}

\end{document}